 \documentclass[preprint,review,12pt]{elsarticle}

\usepackage{graphicx}

\usepackage{amssymb}
\usepackage{dutchcal}

\usepackage{subfig}
\usepackage{tikz}
  \usetikzlibrary{matrix,arrows,decorations.pathmorphing}

\definecolor{gray1}{HTML}{A4A4A4}
\definecolor{gray2}{HTML}{757575}
\definecolor{gray3}{HTML}{FAFAFA}
\definecolor{gray4}{HTML}{F2F2F2}

\tikzstyle{decision} = [diamond, draw, fill=blue!20,
    text width=4.5em, text badly centered, node distance=2.5cm, inner sep=0pt]
\tikzstyle{block} = [rectangle, draw, fill=gray1,
    text width=5em, text centered, rounded corners, minimum height=4em]
\tikzstyle{line} = [draw, very thick, color=black!50, -latex']
\tikzstyle{cloud} = [draw, rectangle,fill=gray4, node distance=2.5cm,
    minimum height=2em, text centered, text width=7em]

\tikzstyle{state}=[shape=circle,double,double distance=1pt,draw=black!50,fill=gray4,minimum size=2cm,inner sep=0pt]
\tikzstyle{start}=[shape=circle,draw=black!50,fill=gray3,minimum size=2cm,inner sep=0pt]
\tikzstyle{symbol}=[shape=circle,draw=white,fill=white,minimum size=1cm,inner sep=0pt]
\tikzstyle{observation}=[shape=rectangle,draw=black!50,fill=gray1,minimum width=1.5cm,minimum height=1cm,inner sep=1pt]
\tikzstyle{lightedge}=[<-,dotted]
\tikzstyle{mainstate}=[state,thick]
\tikzstyle{mainedge}=[<-,thick]

\usepackage{enumitem} 
  \setlist{itemsep=1ex plus0.2ex, leftmargin=*, align=left}
  \setenumerate{label=(\arabic*)}

\makeatletter
\newcommand{\labitem}[2]{%
\def\@itemlabel{\textbf{#1}}
\item
\def\@currentlabel{#1}\label{#2}}
\makeatother

\makeatletter
\newcommand{\headingitem}[1]{%
\vspace{0.3cm}
\def\@itemlabel{\textbf{#1}}
\item
\def\@currentlabel{#1}
\addtocounter{enumi}{-1}
}
\makeatother

\usepackage{csvsimple}
\usepackage{multirow}
\usepackage{lipsum}

\usepackage{eurosym}
\usepackage{siunitx}
  \ifx\ees\undefined
    \DeclareSIUnit\eur{\officialeuro}
    \DeclareSIUnit\M{M}
    \DeclareSIUnit\k{k}
  \else
    \newunit{\eur}{{\mathnormal{\text{\officialeuro}}}}
    \newunit{\M}{M}
    \newunit{\k}{k}
  \fi
\usepackage{csquotes}
\usepackage{amsmath}
\usepackage{xspace}
	\newcommand\ie{i.\,e.\xspace}
	\newcommand\eg{e.\,g.\xspace}

	\newcommand\cf{cf.\xspace}

\usepackage[amsmath,hyperref,framed]{ntheorem} 
  \theoremstyle{plain}
     
  \theoremstyle{nonumberplain}
    \theoremseparator{.}
    \theoremheaderfont{\bfseries}
    \theorembodyfont{\normalfont}
    \theoremsymbol{$\blacksquare$}
      \RequirePackage{amssymb}

      \makeatletter
    \let\copy@theorem@headerfont=\theorem@headerfont
    \newcommand{\my@theorem@headerfont}{%
        \boldmath\copy@theorem@headerfont\unboldmath
      }
    \let\theorem@headerfont=\my@theorem@headerfont
      \makeatother

\theoremstyle{nonumberplain}
\theoremseparator{.}
\usepackage[%
  sort&compress
]{cleveref}
\usepackage{url}

\usepackage{isomath}

\newcommand{\mathup}[1]{\mathrm{#1}}

  \newcommand{\abs}[1]{\left\lvert #1 \right\rvert}

\usepackage{colortbl}
\usepackage{tabularx}
\usepackage{booktabs}
\usepackage{collcell}

\usepackage{ragged2e}
  
\newcommand{\PreserveBackslash}[1]{\let\temp=\\#1\let\\=\temp}
\newcolumntype{v}[1]{>{\PreserveBackslash\RaggedRight\hspace{0pt}}p{#1}}

  \usepackage{adjustbox}
\newcolumntype{Q}[2]{%
    >{\adjustbox{angle=#1,lap=\width-(#2)}\bgroup}%
    l%
    <{\egroup}%
}

\usepackage[
    colorinlistoftodos,
    textsize=footnotesize,
        ]{todonotes}

\newcommand{\cmark}{\ding{51}}%
\newcommand{\xmark}{\ding{55}}%

\makeatletter
    \renewcommand{\fps@figure}{H}         
    \renewcommand{\fps@table}{H}         
\makeatother 

\usepackage{float}
\usepackage{rotating}

\usepackage{pgfplots}
  \pgfplotsset{compat=newest}
\usepackage{pgfplotstable}
  \ifx\ees\undefined
    \usepgfplotslibrary{groupplots}
    \usepgfplotslibrary{dateplot}
    \usepgfplotslibrary{units}
    \usepgfplotslibrary{external}
    \usepgfplotslibrary{groupplots}
  \fi
  \definecolor{VisioDarkBlue}{rgb}{0.4196,0.6078,0.7804}

  \pgfplotscreateplotcyclelist{blue}{%
    {VisioDarkBlue,solid},
    {VisioDarkBlue!40!white,solid},
    {black,solid},
    {Uni-Rot,solid},
    pink,
    magenta,
    RawSienna
  }
  \pgfplotsset{%
    /pgfplots/area cycle list/.style={/pgfplots/cycle list={%
        {violet,fill=violet!60!white},
        {teal,fill=teal!50!white},
        {Blue,fill=Blue!10!white},
        {violet,fill=violet!30!white},
        {teal,fill=teal!30!white},
        {Blue,fill=Blue!30!white},
        {cyan,fill=cyan!30!white},
      }
    },
    /pgfplots/bar cycle list/.style={/pgfplots/cycle list={%
        {VisioDarkBlue,fill=VisioDarkBlue!40!white},
        {VisioDarkBlue,fill=VisioDarkBlue},
        {VisioDarkBlue!60!black,fill=VisioDarkBlue},
        {violet,fill=violet!60!white},
        {teal,fill=teal!40!white},
        {cyan,fill=cyan!30!white},
      }
    },
    legend style={font=\footnotesize},
    label style={font=\footnotesize},
    tick label style={font=\footnotesize},
    every mark/.append style={scale=0.7},
    width=0.95\linewidth,
    height=0.5\linewidth,
    every axis/.append style={line width=0.7pt},
    cycle list name=blue,
    grid=major,
    every axis grid/.append style={dotted,black!40},
    enlarge x limits=0.05,
  }
  
  \ifx\ees\undefined
    \pgfplotsset{
      legend cell align=left,
    }
  \fi

\usepackage[margin=3.5cm]{geometry}
\journal{Journal of Management Information Systems}

\begin{document}

\begin{frontmatter}



\title{Understanding Negations in Information Processing: Learning from Replicating Human Behavior}


\author[Freiburg]{Nicolas Pr\"ollochs\corref{cor1}}
\ead{nicolas.proellochs@is.uni-freiburg.de}

\author[Freiburg]{Stefan Feuerriegel}
\ead{stefan.feuerriegel@is.uni-freiburg.de}

\author[Freiburg]{Dirk Neumann}
\ead{dirk.neumann@is.uni-freiburg.de}

\address[Freiburg]{Chair for Information Systems Research, University of Freiburg, Platz der Alten Synagoge, 79098 Freiburg, Germany}
\cortext[cor1]{Corresponding author. Mail: nicolas.proellochs@is.uni-freiburg.de; Tel: +49\,761\,203\,2395; Fax: +49\,761\,203\,2416.}

\begin{abstract}
Information systems experience an ever-growing volume of unstructured data, particularly in the form of textual materials. This represents a rich source of information from which one can create value for people, organizations and businesses. For instance, recommender systems can benefit from automatically understanding preferences based on user reviews or social media. However, it is difficult for computer programs to correctly infer meaning from narrative content. One major challenge is negations that invert the interpretation of words and sentences. As a remedy, this paper proposes a novel learning strategy to detect negations: we apply reinforcement learning to find a policy that replicates the human perception of negations based on an exogenous response, such as a user rating for reviews. Our method yields several benefits, as it eliminates the former need for expensive and subjective manual labeling in an intermediate stage. Moreover, the inferred policy can be used to derive statistical inferences and implications regarding how humans process and act on negations.
\end{abstract}


\end{frontmatter}



\section{Introduction}
\label{sec:introduction}

When making decisions in their daily lives, humans base their reasoning on information, while pondering expected outcomes and the importance of individual arguments. At the same time, they are continuously confronted with novel information of potentially additional value~\citep{LaBerge.1974}. Psychological theories suggest that, when processing information, humans constantly categorize and filter for relevant tid bits~\citep{LaBerge.1974,Schneider.1977}. The outcome of this filtering then drives decision-making , which in turn affects interactions with information technology, personal relationships, businesses or whole organizations. Information Systems~(IS) research~\citep{Briggs.2015} thus strives for insights into how humans interpret and react to information in order \emph{\textquote{to understand and improve the ways people create value with information}}~\citep[p.\,20]{Nunamaker.2011}. 

Information is increasingly encoded not only in quantitative figures, but also in qualitative formats, such as textual materials~\citep{Lacity.1994}. Common examples from the digital age include blog entries, posts on social media platforms, user-generated reviews or negotiations in electronic commerce. These materials predominantly encompass feedback, comments or reviews, and thus immediately impact the decision-making processes of individuals and organizations~\citep{Chau.2012,Vodanovich.2010}. Rather than a manual text analysis, a computerized evaluation is generally preferable, as it can process massive volumes of documents, often in realtime. Recent advances in information technology render it possible to automatically investigate the influence of qualitative information from narrative language and word-of-mouth -- especially in order to gain an understanding of its content~\citep{Hirschberg.2015}. This, in turn, opens up novel opportunities for computerized decision support, \eg question answering and information retrieval (\eg~\citep{Vlas.2012}). Consequently, understanding decision-making and providing decision support both increasingly rely upon computerized natural language processing. 

The exact interpretation of language is largely impracticable with computer programs at present. Among the numerous difficulties is the particularly daunting challenge of analyzing negations~\citep{Cruz.2015,Pang.2008}, since their context-dependent nature hampers bag-of-words approaches for natural language. The latter method considers only word frequencies without looking at the order of words from the beginning to the end of a document. Such a careful consideration is, however, necessary as negations occur in various forms; they reverse the meanings of individual words, but also of phrases or even whole sentences~\citep{Councill.2010}. Thus, one must handle negations properly in order to achieve an accurate comprehension of texts. The importance of profound language understanding is demonstrated by the following exemplary applications:
\begin{description}
\item[Recommender Systems.] Recommender systems support users by predicting their rating or preference for a product or service. An increasing number of user-generated reviews constitutes a compelling source of information~\citep{Archak.2011}. Hence, recommender systems must accurately classify positive and negative content in order to interpret the intended opinion contained within reviews~\citep{Cruz.2015,Thet.2010,Pang.2008} or their credibility~\citep{Jensen.2013}. 
\item[Financial News.] Investors in financial markets predominantly base their decision on news when choosing whether to exercise stock ownership. In addition to quantitative numbers, qualitative information found in news, such as tone and sentiment, strongly influences stock prices (\eg~\citep{Henry.2008,Tetlock.2007}). For example, companies often frame negative news using positive words~\citep{Loughran.2011}; therefore, empirical research, investors and automated traders demand the precise processing of negations.  
\item[Question Answering Systems.] Question answering systems support users with insights drawn from immense bodies of data. For instance, IBM's Watson processes millions of medical documents in order to discover potential diseases and recommends treatments based on symptoms. Similarly, one can automatically determine software requirements from descriptions. For such applications of information retrieval, it is necessary to distinguish between certainty and beliefs in natural language by considering negations~\citep{CruzDiaz.2012,Rokach.2008,Vlas.2012}. 
\item[Negotiations.] Negotiations usually consists of a seesaw of offers and counter-offers, of which most are usually rejected until one is finally accepted. Even in the digital area, negotiations are based on textual arguments~\citep{Johnson.2015} and, in order for systems to automatically decode outcomes, it is necessary to examine language correctly~\citep{Lai.2002,Twitchell.2013}. Similarly, this holds for cases in which language helps to predict deception in human communication~\citep{Fuller.2013,Zhou.2004}.
\end{description}
Despite the fact that language offers a rich source of information, its processing and the underlying decision-making are still subject to ongoing research activities. This includes negation processing, which affects virtually every context or domain, since neglecting negations can lead to erroneous implications or false interpretations. In the following, we refer to the part of a document whose meaning is reversed as the \emph{negation scope}. Identifying negation scopes is difficult as these are latent, unobservable and -- even among experts -- highly subjective~\citep{Councill.2010}. In addition, many machine learning algorithms struggle with this type of problem as it is virtually impossible to encode with a fixed-length vector while preserving its order and context~\citep{Hirschberg.2015}.

This paper develops a novel method, based on reinforcement learning, for detecting, understanding and interpreting negations in natural language. This approach exhibits several favorable features that overcome shortcomings found in prior works. Among them, reinforcement learning is well suited to learning tasks of varying lengths; that is, it can process sentences of arbitrary complexity while preserving context and order of information. Furthermore, our approach eliminates the need for manual labeling of individual words and thus avoids the detrimental influence of subjectivity and misinterpretation. On the contrary, our model is solely trained on an exogenous response variable at document level. We refer to this score as the \emph{gold standard}, not only  because it represents a common term in text mining, but also to stress that it can reflect any human behavior of interest. As a result of its learning process, our method adapts to domain-specific cues or particularities of the given prose. 


Our contribution goes beyond the pure algorithmic benefits, since we envision the goal of understanding human information processing. For this purpose, our approach essentially learns to replicate human decision-making in order to gain insights into the workings of the human mind when processing narrative content. In fact, the reinforcement learning approach is trained based on past decisions. While the model receives feedback as to how well it matches the human decision, it does not receive explicit information regarding how to improve accuracy. Rather, the learner iteratively processes information in textual materials and experiments with different variants of negation processing in a trial-and-error manner that imitates human behavior. 

Reinforcement learning can considerably advance our understanding of decision-making. Indeed, learning itself has long been conceptualized as the process of creating new associations between stimuli, actions and outcomes, which then guide decision-making in the presence of similar stimuli~\citep{Niv.2015}. We thus show how to make the knowledge of these associations explicit: we propose a framework by which to study negation scopes that were previously assumed to be latent and unobservable. Contrary to this presumption, we manage to measure their objective perception. Therefore, we exploit the action-value function inside the reinforcement learning model in order draw statistical inferences and derive conclusions regarding how the human mind processes and acts upon negations. As such, our approach presents an alternative or supplement to experiments, such as those of a psychological or neuro-physiological nature (along the lines of NeuroIS). 

This paper is structured as follows. \Cref{sec:related_work} provides an overview of related works that investigate human information processing of textual materials, while also explaining the motivation behind our research objective of understanding negations in natural language. Subsequently, \Cref{sec:methodology} explains how we adapt reinforcement learning to improve existing methods of negation scope detection. In \Cref{sec:evaluation}, we demonstrate our novel approach with applications from recommender systems and finance in order to contribute to existing knowledge of information processing. Finally, \Cref{sec:discussion} discusses implications for IS research, practice and management.

\section{Background}
\label{sec:related_work}

This section presents background on natural language processing. First, we discuss recent advances in computational intelligence and then outline challenges that arise when working with narrative content. We conclude by briefly reviewing previous works on the handling of negation in natural language.  

\subsection{Human Information Processing}

Advances in computational intelligence have revolutionized our understanding of learning processes in the human brain. As a result, research has yielded precise theories regarding the reception of information and function of human memory~\citep{Niv.2015}. For instance, statistical models provide insights into human memory formation and the dynamics of memory updating, while also validating these theories by replicating experiments with statistical computations~\citep{Gershman.2014}. Reinforcement learning, especially, has gained considerable traction as it mines real experiences with the help of trial-and-error learning to understand decision-making~\citep{Niv.2015}. Accordingly, existing studies find that the brain naturally reduces the dimensionality of real-world problems to only those dimensions that are relevant for predicting the outcome~\citep{Niv.2015}. Along these lines, a recent review argues for jointly combining both perception and learning in order to draw statistical inferences regarding information processing~\citep{Fiser.2010}. While the previous reference materials predominantly address visual perception, the focus of this paper is rather on natural language. 

Behavioral theories suggest that human decision-makers seek as much information as possible in order to make an informed decision~\citep{Wilson.1999}. In the case of natural language, researchers have devised advanced methods to study the influence of textual information on the resulting decision. On the one hand, it is common to extract specific facts or features from the content and relate these to a decision variable~\citep{Thet.2010}. On the other hand, information diffusion is also frequently studied by measuring the overall tone of documents. This latter approach comprises a variety of different aspects of perception, including negative language, sentiment and emotions~\citep{Pang.2008,Stieglitz.2013,Tetlock.2007}.

In the case of natural language, a variety of textual sources have served as research subjects for studying word-of-mouth communication and information diffusion. For instance, the dissemination of information and sentiment has been empirically tested in social networks~\citep{Trung.2014}, revealing that emotionally-charged tweets are retweeted more often and faster~\citep{Stieglitz.2013}. Similarly, measuring the response to information allows one to test behavioral theories, such as attribution theories or the negativity bias, by distinguishing between the reaction to positive and negative content~\citep{Aggarwal.2012,Stieglitz.2013}. 


\subsection{Natural Language Processing}

Addressing the above research questions on information processing requires accurate models for understanding and interpreting natural language. However, the majority of such methods only count the occurrences of words (or combinations), resulting in so-called \emph{bag-of-words} methods. By doing so, these techniques have a tendency to ignore information relating to the order of words and their context~\citep{Hirschberg.2015}, such as inverted meanings through negations. 

Neglecting negations can substantially impair accuracy when studying human information processing; for example, it is common \emph{\textquote{to see the framing of negative news using positive words}}~\citep{Loughran.2011}. To avoid false attributions, one must identify and predict negated text fragments precisely, since information is otherwise likely to be classified erroneously. This holds true not only for negations in information retrieval~\citep{CruzDiaz.2012,Rokach.2008}, but especially when studying sentiment~\citep{Cruz.2015,Wiegand.2010}; even simple heuristics can yield substantial improvements in such cases~\citep{Jia.2009}. 


\subsection{Negation Processing}

Previous methods for detecting, handling and interpreting negations can be grouped into different categories (\cf~\citep{HICSS.2015,HICSS.2016,Rokach.2008}). 

Rule-based approaches are among the most common due to their ease of implementation and solid out-of-the-box performance. In addition, rules have been found to work effectively across different domains and rarely need fine-tuning~\citep{Taboada.2011}. They identify negations based on pre-defined lists of negating cues and then hypothesize a language model which assumes a specific interpretation by the audience. For example, some rules invert the meaning of all words in a sentence, while others suppose a forward influence of negation cues and thus invert only a fixed number of subsequent words~\citep{Hogenboom.2011}. Furthermore, a rule-based approach can also incorporate syntactic information in order to imitate subject and object~\citep{Padmaja.2014}. However, rules cannot effectively cope with implicit expressions or particular, domain-specific characteristics.

Machine learning approaches can partially overcome previous shortcomings~\citep{Rokach.2008}, such as the difficulty of recognizing implicit negations. Common examples of such methods include generative probabilistic models in the form of Hidden Markov models and conditional random fields (\eg~\citep{Councill.2010}). These methods can adapt to domain-specific language, but require more computational resources and rely upon ex ante transition probabilities. Although approaches based on unsupervised learning avoid the need for any labels, practical applications reveal inferior performance compared to supervised approaches~\citep{HICSS.2015}. The latter usually depend on manual labels at a granular level, which are not only costly but suffer from subjective interpretations~\citep{Councill.2010}. 




\section{Method Development}
\label{sec:methodology}

This section posits the importance of developing a novel method for learning negation scopes in textual materials. After first formulating a problem statement, we introduce our approach, which is based on reinforcement learning.

\subsection{Rationale and Intuition of Proposed Methodology}


Negation scope detection in related research predominantly relies on rule-based algorithms. 
Rule-based approaches entail several drawbacks, as the list of negations must be pre-defined and the selection criterion according to which rule a rule is chosen is usually random or determined via cross validation. Rules aim to reflect the \textquote{ground truth} but fail at actually learning this. 

\begin{figure}
\centering
    \includegraphics[width=\linewidth]{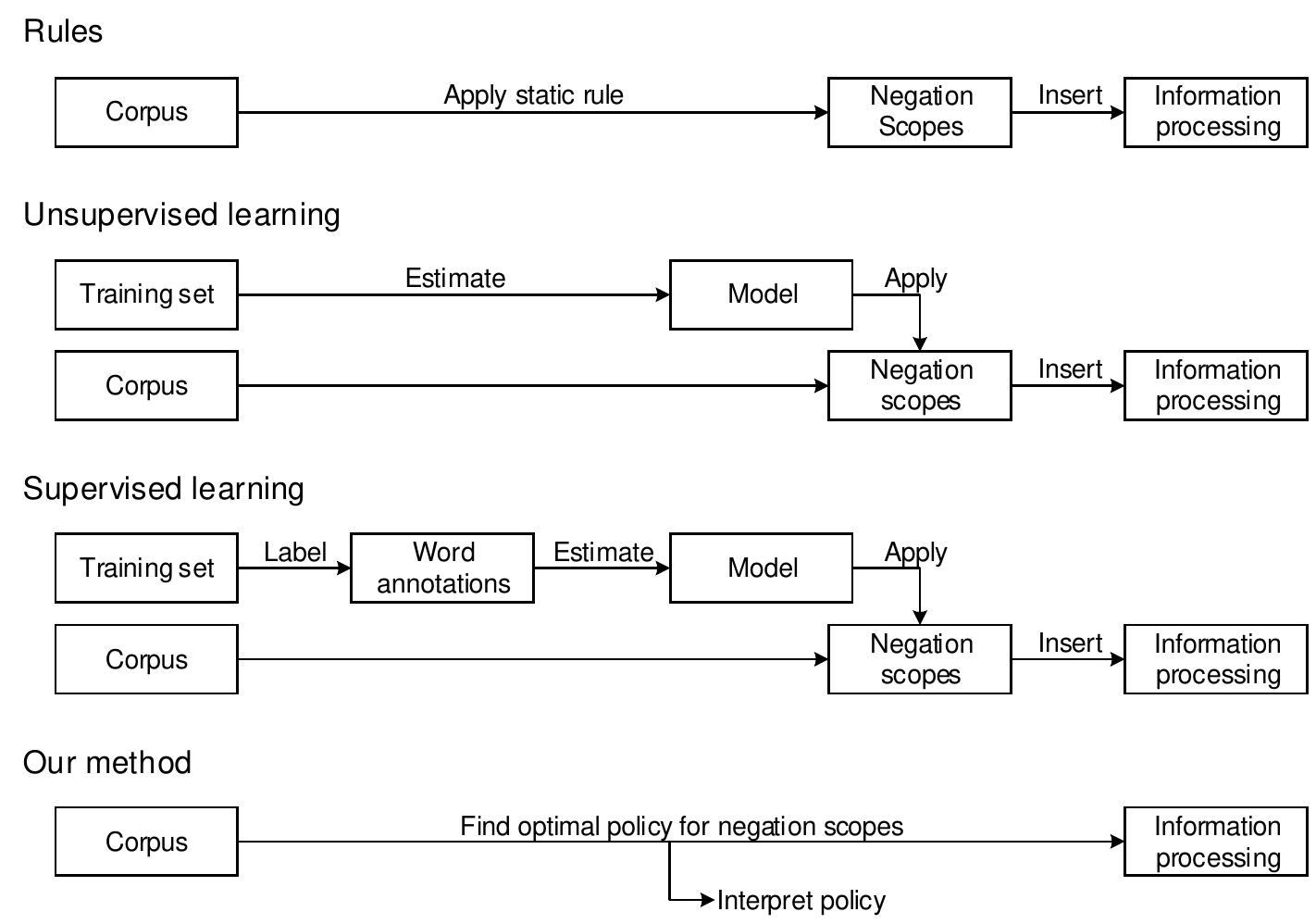}
\caption{Process chart compares the different stages of labeling, training and applying rules in order to evaluate information processing.}
\label{fig:learning_process}
\end{figure}

For those seeking to incorporate a learning strategy, a viable alternative exists in the form of generative probabilistic models (\eg Hidden Markov models or conditional random fields~\citep{Rokach.2008}). These process narrative language word-by-word and move between hidden states representing negated and non-negated parts. On the one hand, unsupervised learning can estimate the models without annotations, but yields less accurate results overall~\citep{HICSS.2015}. On the other hand, supervised learning offers better performance, but this approach requires a training set with manual labels for each word (see \Cref{fig:learning_process}) which are supposed to approximate the latent negation scopes. Such labeling requires extensive manual work and is highly subjective, thus yielding only fair performance~\citep{HICSS.2015,HICSS.2016}. As a further drawback, many approaches from supervised machine learning are simply infeasible as they usually require an input vector of a fixed, pre-defined length without considering its order. This circumstance thus necessitates a tailored method for dealing with negations in narrative materials.

In contrast to these suboptimal methods, we propose a novel approach to determining latent negation scopes based on reinforcement learning. It works well with learning tasks of arbitrary length~\citep{Sutton.1998} and is adaptable to domain-specific features and particularities. Since it relies only upon a gold standard at document level, it represents a more objective strategy. However, such an approach has been largely overlooked in previous works on natural language processing (see \Cref{sec:related_work}).

Reinforcement learning aims at learning a suitable policy directly through trial-and-error experience. It updates its knowledge episodically and learns the policy from past experience using only limited feedback in the form of a reward. This reward indicates the current performance of the classifier, but does not necessarily specify how to improve the policy. In addition, this type of learning can also handle highly complex sentences and is thus well suited to the given task. 

\subsection{Learning Task for Negation Detection}

Understanding negations and their influence on language is -- as previously mentioned -- a non-trivial computational problem, since the underlying learning task suffers from several undesirable features: 
\begin{enumerate}
\item Even though sentences follow grammatical rules, they can be nested up to arbitrary complexity and thus become arbitrarily long. 
\item Words have a meaning based on their context, which is implicitly established by their order. By merely rearranging the word order, one can produce a completely different meaning. This constitutes a dependency according to which the meaning of words depends, in part, on all other words and their order in the same document. 
\item Negation scopes affect individual words; however, we lack annotations on a word-by-word basis. Instead, we only observe a gold standard for the whole document upon which we must reconstruct negation scopes for each individual word within the document. 
\end{enumerate}
Based on these challenging features, we can formalize the problem, resulting in the following learning task. Both negation scopes and sentences are of different length $N_d$ depending on the specific document $d$. This length can theoretically range from one to infinity. Each word $w_{d,i}$ in document $d$ with $i \in \{ 1, \ldots, N_d \}$ thus represents an individual classification task, which also depends on all other words in that document, \ie
\begin{equation}
f : ( w_{d,i}, \underbrace{\left[ w_{d,1}, w_{d,2}, \ldots, w_{d,N_d} \right]}_{\text{Ordered sequence of words}} ) \mapsto \{ \text{Negated}, \neg\text{Negated}\} ,
\end{equation}
where $\left[ w_{d,1}, w_{d,2}, \ldots \right]$ is an ordered list of variable length providing context information. 

Each document comes with a single label $y_{d}$, \ie the gold standard, which reflects the response of human decision-making to the text processing. In order to estimate $f$, we minimize the expected error (or any other loss-like function) via the gold standard and the result of a text processing function. The latter function maps the words as a predictor onto the gold standard. Examples of text processing functions are functions that measure the accuracy of information retrieval or sentiment based on the presence of polarity words. 

\subsection{Reinforcement Learning}

Reinforcement learning constructs a suitable policy for negation classification through trial-and-error experience. That is, it mimics human-like learning and thus appears well suited to natural language processing. In the following section, we introduce its key elements and tailor the method to our problem statement.

The overall goal is to train an \emph{agent} based on a recurrent sequence of interactions. After observing the current state, the agent decides upon an action. Based on the result of the action, the agent receives immediate feedback via a reward. It is important to note that the agent aims only to maximize the rewards, but it never requires pairs of input and the true output (\ie words and a flag indicating whether they are negated). This forms a setting in which the agent learns the latent negation scopes. 

\begin{figure}
\centering
    \includegraphics[width=.6\linewidth]{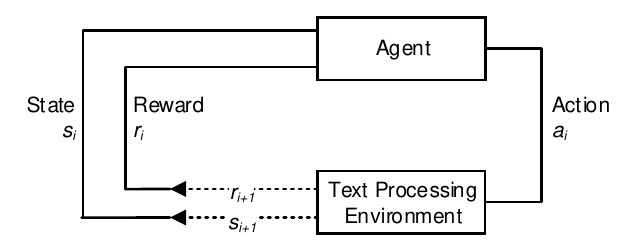}
\caption{Interaction between agent and environment in reinforcement learning~\citep{Sutton.1998}.}
\label{fig:agent_environment}
\end{figure}

More formally, the model consists of a finite set of environmental states $S$ and a finite set of actions $A$. The agent models the decision-maker by iteratively interacting with an environment over a sequence of discrete steps and seeks to maximize the reward over time. Here, the \emph{environment} is a synonym for the states and the transition rules between them. At each iteration $i$, the decision-making agent observes a \emph{state} $s_i \in S$. Based on the current state $s_i$, the agent picks an action $a_i \in A(s_i)$, where $A(s_i) \subseteq A$ is the subset of available actions in the given state $s_i$. Subsequently, the agent receives feedback related to its decision in the form of a numerical reward $r_{i+1}$, after which it moves then moving to the next state $s_{i+1}$. The entire process is depicted in \Cref{fig:agent_environment}.

In order to build up knowledge, reinforcement learning updates a \emph{state-action function} $Q(s_i, a_i)$, which specifies the expected reward for each possible action $a_i$ in state $s_i$. This knowledge can then be used to infer the optimal behavior, \ie the policy that maximizes the expected reward from any state. Specifically, the \emph{optimal policy} $\pi^*(s_i,a_i)$ chooses in state $s_i$ the actions $a_i$ that maximizes $Q(s_i,a_i)$. 

%
		%

Several algorithms have been devised to learn an optimal policy $\pi^\ast$, among which is an approach known as Q-learning~\citep{Sutton.1998,Watkins.1992}. This methods seeks an optimal policy without an explicit model of the environment. In other words, it knows neither explicitly the reward function nor the state transition function~\citep{Hu.2003}. Instead, it iteratively updates its action-value $Q(s_t,a_t)$ based on past experience~\citep{Watkins.1992}. In our case, we use a variant with eligibility traces named Watkin's $Q(\lambda)$ due to better convergence; see~\citep{Sutton.1998} for details.

\subsection{Learning Negation Processing}

In this section, we outline how we adapt reinforcement learning to our attempt to simulate human negation processing. In each iteration, the agent observes the current state $s_i = ( w_i , a_{i-1} )$ that we engineer as the combination of the $i$-th word $w_i$ in a document and the previous action $a_{i-1}$. This specification establishes a recurrent architecture whereby the previous negation can pass on to the next word.\footnote{Such a design is common in partially observable Markov decision processes (POMDP for short) that feature a similar relaxation into so-called belief states~\citep{Kaelbling.1996}.} At the same time, this allows for nested negations, as a word can first introduce a negation scope and another subsequent negation can potentially revert it based on $a_{i-1} = \text{Negated}$ to follow a non-negating action again. In our case, we incorporate the actual words into the states, while other variants are also possible, such as using part-of-speech tags or word stems instead. The latter variants work similarly; however, our tests suggest a lower out-of-sample performance.\footnote{By definition, the use of $n$-grams is not necessary, as the context is implicitly modeled by the ordered sequence of states and actions.}

After observing the current state, the agent chooses an action $a_t$ from of two possibilities: (1)~it can set the current word to \emph{negated} or (2)~it can mark it as \emph{not negated}. Hence, we obtain the following set of possible actions $
A = \{ \text{Negated}, \neg\text{Negated} \}
$.
Based on the selected action, the agent receives a reward, $r_i$ which updates the knowledge in the state-action function $Q(s_i,a_i)$. This state-action function is then used to infer the best possible action $a_i$ in each state $s_i$, \ie the optimal policy $\pi^*(s_i,a_i)$.

Our approach relies upon a text processing function that measures the correlation between a given gold standard at document level (\eg the author's rating in movie reviews) and the content of a document. We later show possible extensions (see \Cref{sec:extensibility}), but for now demonstrate only how the tone (or sentiment) in a document works as a predictor of its exogenous assessment. Examples of such predicted variables are movie ratings in the case of reviews or stock market returns in the case of financial disclosures. Even though more advanced approaches from machine learning are possible, we prefer -- for reasons of clarity -- an approach is based on pre-defined lists of positive and negative terms, $L_{\mathup{pos}}$ and $L_{\mathup{neg}}$. We then measure the tone $S_d$ in document $d$ as the difference between positively and negatively opinionated terms divided by the overall number of terms in that document~\citep{Pang.2008}, \ie
\begin{equation}
S_d = \frac{\abs{\left\{ w_i \,|\, w_i \in L_{\mathup{pos}} \right\}} - \abs{\left\{ w_i \,|\, w_i \in L_{\mathup{neg}} \right\}}}{N_d} .
\end{equation}
 If a term is negated by the policy, the polarity of the corresponding term is inverted, \ie positively opinionated terms are counted as negative and vice versa. 
 Our list of opinionated words originates from the Harvard~IV General Inquirer dictionary which contains \num{1915} opinionated entries with a positive connotation and \num{2291} entries marked as negative.

Let us now demonstrate the learning process via an example, where the agent processes word-by-word the first document, which consists of \emph{\textquote{this is a good product}}, with gold standard $+1$ (\ie positive content). The agent might then, at random, decide to explore the environment by negating the word \textquote{\emph{good}}. Upon reaching the last word, it receives feedback in the form of a zero as there is no improvement from having negation scopes compared to having none. Thus, the agent will discard this action in the future. It then processes the second document: (\emph{\textquote{this isn't a good product}} with gold standard $-1$), where it negates all words following \textquote{\emph{isn't}}. As this inversion now better reflects the gold standard, the agent receives a positive reward and will apply this rule in the future. Ultimately, the agent is also able to learn a suitable policy for nested negations, \eg for the third document \emph{\textquote{this product isn't good but fantastic}} with gold standard $+1$. Based on the current policy, it negates all words following \textquote{\emph{isn't}} but receives a negative reward as there is an inferior resulting correlation compared to not incorporating negations. However, through further exploration, the agent learns that it is beneficial to terminate the negation scope after \textquote{\emph{but.}} Thus, the agent will invert all words subsequent to \textquote{\emph{isn't}} and terminate the negation scope subsequent to \textquote{\emph{but}} if (and potentially only if) the previous state is negated. \Cref{tbl:qFunctionNegations} illustrates an exemplary resulting state-action function for this learning process.

\begin{table}
\sisetup{round-mode=places,round-precision=4}
\vspace{0pt}
\centering
\renewcommand{\arraystretch}{1.3}
\tiny
	\begin{tabular}{lSS|r}
		\toprule
			\textbf{State=\boldmath{$( w_i , a_{i-1} )$}} & \boldmath{$a_i=\textbf{Negated}$} & \boldmath{$a_i=\neg\textbf{Negated}$} & \boldmath{$\pi^*(s_i,a_i)$} \\
			\midrule
			$({\textit{this}},\neg{\text{Negated}})$  & 3 &  6 & $\neg{\text{Negated}}$\\
			$({\textit{product}},\neg{\text{Negated}})$ & 1 &  3 & $\neg{\text{Negated}}$\\
			$({\textit{isn't}},\neg{\text{Negated}})$ & 5 &  1 & ${\text{Negated}}$\\
			$({\textit{good}},{\text{Negated}})$ & 3 &  1 & ${\text{Negated}}$\\
			$({\textit{but}},{\text{Negated}})$ & 2 &  4 & $\neg{\text{Negated}}$\\
			$({\textit{fantastic}},\neg{\text{Negated}})$ & 2 & 3 & $\neg{\text{Negated}}$\\

		
			\bottomrule
	\end{tabular}
\caption{Exemplary table for the state-action function \boldmath{$Q(s_i,a_i)$} with recurrent state architecture and actions \boldmath{$A = \{ \text{Negated}, \neg\text{Negated} \}$}. Each cell contains the expected reward for the corresponding state-action pairs. The last column shows optimal policy \boldmath{$\pi^*(s_i,a_i)$} for all states and actions.}
\label{tbl:qFunctionNegations}
\end{table}

We now specify a reward $r_i$ such that it incentivizes the outcome of the text processing function to match the gold standard. When processing a document, we cannot actually compute the reward (as not all negations are clear) until we have processed all words. Therefore, we set the reward before the last word to almost zero, \ie $r_i \approx 0$ for all $i = 1, \ldots, N_d - 1$. Upon reaching the final word, the agent compares the text processing function without any negation to the current policy $\pi^\ast$. The former is defined by the absolute difference between gold standard $y_d$ and tone $S_d^0$, whereas the latter is defined by the absolute difference between gold standard $y_d$ and the adjusted tone using the current policy $S_d^\pi$. Then the difference between the text processing functions returns the terminal reward $r_{N_d}$. This results in the reward
\begin{equation}
r_i = \begin{cases}
0 , & \text{if } a_i = \text{Negated and } i < N_d , \\
c , & \text{if } a_i = \neg\text{Negated and } i < N_d , \\
\abs{y_d - S_d^0} - \abs{y_d - S_d^\pi} , & \text{if } i = N_d,
\end{cases}
\end{equation}
with constant $c=0.005$ that adds a small reward for default (\ie non-negating) actions to avoid overfitting.

At the beginning, we initialize the action-value function $Q(s,a)$, \ie the current knowledge of the agent, to zero for all states and actions.\footnote{This also controls our default action when encountering unknown states or words in the out-of-sample dataset. In such cases, the non-negated action is preferred.} 
The agent then successively observes a sequence of words in which it can select between exploring new actions or taking the current optimal one. This choice is made by $\varepsilon$-greedy selection according to which the agent explores the environment by selecting a random action with probability $\varepsilon$ or, alternatively, exploits the current knowledge with probability $1-\varepsilon$. In the latter case, the agent chooses the action with the highest estimated reward for the given policy.\footnote{First, we perform \num{4000} iterations with a higher exploration rate as given by the following parameters: exploration $\varepsilon = \SI{0.1}{\percent}$, discount factor $\gamma = \SI{0}{\percent}$ and learning rate $\alpha = \SI{0.5}{\percent}$. In a second phase, we run \num{1000} iterations for fine-tuning with exploration $\varepsilon = \SI{0.01}{\percent}$, discount factor $\gamma = \SI{0}{\percent}$ and learning rate $\alpha = \SI{0.1}{\percent}$. See~\citep{Sutton.1998,Watkins.1992} for detailed explanations.}

\subsection{Inferences, Understanding and Hypothesis Testing}

Our approach features several beneficial characteristics that make inferences and statistical testing easy. In contrast to black-box approaches in machine learning, we can use the state-action function $Q(s,a)$ to infer rules regarding how the content is processed because the function reflects the ground truth. For instance, this state-action function specifically determines which cues introduce explicit or implicit negations. Additionally, we can gain a metric of confidence about the rules by comparing the largest reward to all other rewards in a specific state. A larger discrepancy expresses higher confidence with regard to a certain action. 

Applying the policy to out-of-sample documents benchmarks its performance in comparison to the absence of negation handling. Furthermore, we can study, for instance, which cues prompt this policy to introduce negation scopes, as well as their position, size or other characteristics as a basis for statistical testing. 

\section{Evaluating Negation Processing}
\label{sec:evaluation}

This section evaluates our method for replicating human negation processing. First, we show how policy learning can help to yield a more accurate interpretation of movie reviews. We then detail the role of negation cues and compare explicit versus implicit negations. In the next step, we validate the robustness of our results and introduce a second application scenario which addresses the relevance of accurate negation handling in financial disclosures. 

\subsection{Case Study: Recommender System}

Recommender systems can benefit greatly from user-generated reviews, which represent a rich source of information. We thus demonstrate our method using a common dataset~\citep{Pang.2005} of \num{5006} movie reviews from the Internet Movie Database archive (IMDb), each annotated with an overall rating at document level.\footnote{We use the scaled dataset available from \url{www.cs.cornell.edu/people/pabo/movie-review-data/}. All reviews are written by four different authors and preprocessed, \eg by removing explicit rating indicators~\citep{Pang.2005}.}
It is widely accepted that measuring the tone of movie reviews is particularly difficult because positive movie reviews often mention some unpleasant scenes, while negative reviews, conversely, often detail certain pleasant scenes~\citep{Turney.2002}. Thus, this corpus appears particularly suitable for a case study since it allows one to examine the importance of human-like negation processing beyond simple rules. Accordingly, we use 10-fold cross validation to verify the predictive accuracy.

\subsection{Policy Learning for Negation Processing}

Shown below are the results from policy learning for negation processing. \Cref{tbl:performance} summarizes the main results and we explicate these findings in more depth. As part of a benchmark, we study the proportion of variance of the gold standard that is explained by the tone when leaving negations untreated. We observe an in-sample $R^2$ of \num{0.0870} and a $R^2$ of \num{0.0867} in the out-of-sample set. We then compare this to our approach of policy learning. We perform \num{5000} learning iterations, thereby yielding significant improvements: the in-sample $R^2$ increases by \SI{158.94}{\percent}, leading to an overall $R^2$ of \num{0.2233}. Similarly, we see a rise by \SI{58.94}{\percent} to a $R^2$ of \num{0.1378} in the out-of-sample set. A better handling of negations thus contributes to more accurate text processing (we later perform additional robustness checks; see \Cref{sec:robustness_checks}).

\begin{table}
\vspace{0pt}
\centering
\sisetup{round-mode=places,round-precision=4}
\renewcommand{\arraystretch}{1.3}
\tiny
	\begin{tabular}{lS[table-format=0.4,round-precision=4]S[table-format=0.4,round-precision=4]S[table-format=2.2,round-precision=2]}
		\toprule
			\textbf{} & \textbf{\renewcommand{\arraystretch}{0.6}\begin{tabular}{c}{\boldmath{$R^2$}}\\(no negation handling)
			\end{tabular}} & \textbf{\renewcommand{\arraystretch}{0.6}\begin{tabular}{c}{\boldmath{$R^2$}}\\(with negation policy)
			\end{tabular}} & \textbf{\renewcommand{\arraystretch}{0.6}\begin{tabular}{c}Improvement\\(in \%)\\
			\end{tabular}}\\
			\midrule
			In-sample set& 0.08699338 & 0.2232798  & 158.94\\
			Out-of-sample set & 0.08668691 & 0.1377806  & 58.94\\
			\bottomrule
	\end{tabular}
\caption{Comparison of gold standard variance explained by the tone. Figures are reported for both the in-sample and out-of-sample sets using 10-fold cross validation after \num{5000} learning iterations.}
\label{tbl:performance}
\end{table}

We now provide descriptive statistics of negation scopes in order to gain further insights (\Cref{tbl:descriptivesOOS}). For this purpose, we apply the learned policy to the out-of-sample documents and record its effects. In the first place, the policy negates a large share (\SI{38.11}{\percent}) of opinionated words, \ie words that convey a positive or negative polarity. On average, each document contains \num{75.18} separate negation scopes of different size and extent, all of which invert opinion words. For example, the length of the corresponding negation scopes, \ie sequences that are uniformly negated, is unevenly distributed and ranges from \num{1} to \num{18} words, whereas the average length of each scope is \num{1.74} words. \SI{75.46}{\percent} of all negation scopes consist of only a single word, while \SI{24.53}{\percent} encompass two or more words.

\textbf{\begin{table}
\sisetup{round-mode=places,round-precision=2}
\vspace{0pt}
\centering
\renewcommand{\arraystretch}{1.3}
\tiny
	\begin{tabular}{lr}
		\toprule
			Minimum length of negation scopes & 1\\
			Maximum length of negation scopes & 18\\
			Mean length of negation scopes & 1.74\\
			Share of negation scopes with $\leq 1$ word &  \SI{75.46}{\percent}\\
			Share of negation scopes with $\geq 2$ word &  \SI{24.53}{\percent}\\
			Share of negated polarity words &  \SI{38.11}{\percent}\\
Mean number of negation scopes per document & {75.18}\\
			\bottomrule
	\end{tabular}
\caption{Descriptive statistics on the out-of-sample set after applying the in-sample policy.}
\label{tbl:descriptivesOOS}
\end{table}
}

\subsection{Negation Cues}

Negation scopes are typically initiated by specific cues that invert the meaning of surrounding words. These negation cues can be grouped in two categories. On the one hand, a negation cue can be explicit, such as \emph{not} in the sentence, \emph{\enquote{This is not a terrible movie}}. On the other hand, negations can also flip the meaning of sentences implicitly, \eg \emph{\enquote{The actor did a great job in his last movie; it was the first and last time}}. 

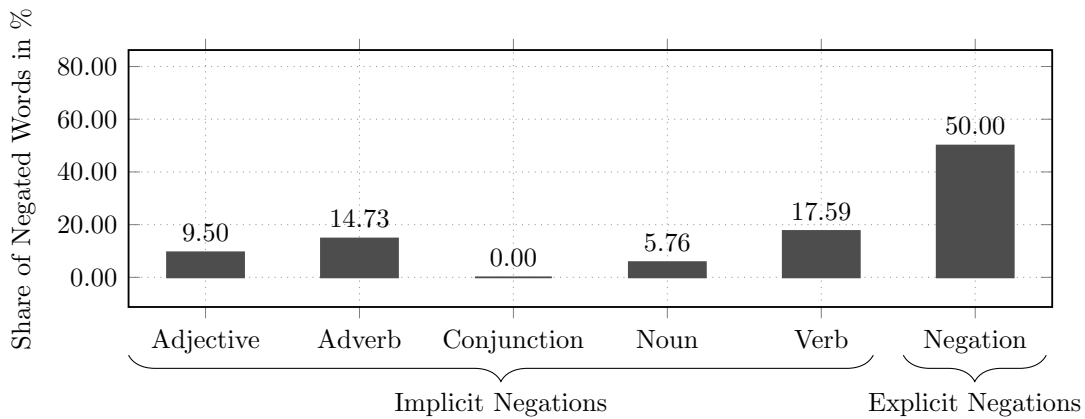
\begin{figure}[ht]
\centering
  \begin{tikzpicture}
    \begin{axis}[
        ybar,
        bar width=0.5,
        ylabel=Share of Negated Words in \%,
        xtick=data,
        xticklabels={Adjective,Adverb,Conjunction,Noun,Verb,Negation},
        yticklabel style={
          /pgf/number format/.cd,
          /pgf/number format/fixed,
          /pgf/number format/fixed zerofill,
          /pgf/number format/precision=2},
        ymax=75,
        enlargelimits=0.15,
        enlarge x limits=0.1,
        height=5cm,
        xlabel=,
        legend pos=south west,
        nodes near coords,
        every node near coord/.style={
          font=\footnotesize,
          /pgf/number format/fixed,
          /pgf/number format/fixed zerofill,
          /pgf/number format/precision=2,
        },
        nodes near coords align={vertical},
      ]
      \addplot[black!70,fill=black!70,text=black] table[col sep=comma,header=false,x index=0,y index=1] 
      {
1,9.495871
2,14.733542
3,0.0000
4,5.762370
5,17.593720
6,50.0000
      };
    \end{axis}
		\draw [decorate,decoration={brace,amplitude=10pt,mirror,raise=4pt},yshift=0pt]
		(0.0,-0.5) -- (9.9,-0.5) node [black,midway,xshift=-0.0cm,yshift=-0.8cm] 
{\footnotesize Implicit Negations};
		\draw [decorate,decoration={brace,amplitude=10pt,mirror,raise=4pt},yshift=0pt]
		(10.3,-0.5) -- (12.2,-0.5) node [black,midway,xshift=-0.0cm,yshift=-0.8cm] 
{\footnotesize Explicit Negations};
  \end{tikzpicture}
\caption{Negation cues per word class based on policy learned after \num{5000} iterations.}
\label{fig:shareNegatedPerWordClass}
\end{figure}

Given this understanding, we investigate individual effects of explicit and implicit negations on text reception. For this purpose, we group the words that initiate a negation scope according to their part-of-speech tag and depict in \Cref{fig:shareNegatedPerWordClass} the resulting share of negation cues by word class. Here, the last bar relies on a list of explicit negations as proposed by~\citep{Jia.2009}. We thus find evidence that a major share of negations (4 out of 8, \ie \SI{50.00}{\percent}) are evoked by explicit cues, while the remainder originate from implicit negations.

We now provide additional descriptions of the appearance of implicit negations. For instance, we frequently observe words such as \emph{fairly} or \emph{hopefully} as part of implicit negation cues, which often transform a positive statement into a negative one, \eg \emph{\enquote{Hopefully, the movie is better next time}}. Contrary to our prior expectations, conjunctions seem unlikely to initiate a negation scope, but are often accountable for double negations. Here, words such as \emph{but} and \emph{nor} frequently revert the meaning of negated words, \ie terminate the negation scope, as in the sentence, \emph{\enquote{The movie is not great but absolutely unmissable}}.

\Cref{tbl:topNegations} provides statistics for all explicit negation cues based on~\citep{Jia.2009}.\footnote{Here, we divide the corpus into two subsets: an in-sample set of \num{4005} reviews which we use to learn the agent, and (b)~an out-of-sample set with the remaining \num{1001} documents to test implications. We exclude the use of cross validation since we desire a single model with which can perform statistical analyses.} Their frequency in the documents differs considerably, while some cues also involve a larger negation scope than others. For example, the negation word \emph{not} negates \num{1.60} subsequent words on average, whereas this figure stands at \num{1.92} for the term \emph{without}. Based on the $Q$-value, we assess their strength, \ie a larger value indicates a higher reward from negating. We can also gain confidence in negations by comparing the gap between the highest and second highest $Q$-value of each word. Interestingly, several words that were previously considered negation cues do not negate surrounding words in our case, namely, \emph{barely}, \emph{less}, \emph{hardly} and \emph{rarely}. 

\begin{table}
\sisetup{round-mode=places,round-precision=4}
\vspace{0pt}
\centering
\renewcommand{\arraystretch}{1.3}
\tiny
	\begin{tabular}{lcS[table-format=1.4,round-precision=4]S[table-format=1.4,round-precision=4]S[table-format=4.0,round-precision=0]S[table-format=1.2,round-precision=2]}
		\toprule
			\textbf{Word} & \textbf{Negating Action} & \textbf{Q-Value} & \textbf{\renewcommand{\arraystretch}{0.6}\begin{tabular}{@{}c@{}}Confidence~(Difference\\to Second Best Policy)\\\end{tabular}} & \textbf{Occurrences} & \textbf{\renewcommand{\arraystretch}{0.6}\begin{tabular}{@{}c@{}}Mean Length\\of Negation Scope\\\end{tabular}}\\
			\midrule
			not & \cmark & 0.07004141 & 0.045588346 & 1941 & 1.599691\\
			no & \cmark & 0.06960190 & 0.049061069 & 699 & 1.705293\\
			never & \cmark & 0.06800171 & 0.036654995 & 398 & 1.673367\\
			without & \cmark & 0.07794868  & 0.041899437 & 249 & 1.915663 \\
			barely & \xmark & {--} & {--} & {--} & {--} \\
			less & \xmark & {--} & {--} & {--} & {--} \\
			hardly & \xmark & {--} & {--} & {--} & {--} \\
			rarely & \xmark & {--}  & {--} & {--} & {--} \\
			\bottomrule
	\end{tabular}
\caption{Explicit negation words from~\citep{Jia.2009} according to policy learned after \num{5000} iterations.}
\label{tbl:topNegations}
\end{table}

This provides evidence that static negation lists are generally inadequate in mimicking human perception. Even though explicit negations can be recognized with predefined lists of cues, implicit ones are often hidden and difficult to identify algorithmically. As a remedy to this shortcoming, our approach is capable of learning both kinds of negations and can handle them accordingly. 


\subsection{Behavioral Implications of Negation Processing}

Policy learning is also a valuable tool for analyzing behavioral implications. In this section, we demonstrate potential policy learning applications that allow for the testing of certain hypotheses regarding human information processing of natural language. 


As an example, our method allows one to test the hypothesis whether negations appear evenly throughout different parts of narrative content. Such a test is not tractable for rules or supervised learning with intermediate labeling, since these introduce a subjective choice of negation cues, rules or labels; however, our method infers a negation policy model from an exogenous response variable. Hence, we can evaluate where authors place negations when composing reviews, \ie do they generally introduce negative aspects in the beginning or rather at the end? We thus compare the frequency of negations (as a proxy for negativity) across different parts of documents in the corpus. Let $\mu_1$ denote the mean of negated words in the first half and $\mu_2$ in the second half of a document or sentence, respectively. We can then test a null hypothesis $H_0: \mu_1 = \mu_2$ to infer behavioral implications. 


As a result, we find that the second half of an out-of-sample document contains \SI{1.38}{\percent} more negations than its first half on average. This difference is statistically significant at even the \SI{0.1}{\percent} significance level when performing a two-sided Welch $t$-test. It also coincides with psychological research according to which senders of information are more likely to place negative content at the end~\citep{Legg.2014}; however, we can provide evidence outside of a laboratory setting by utilizing human information behavior in a real-life environment. Furthermore, the share of negated words also varies across different segments of sentences. However, at this level, the effect tends in the opposite direction as the first half of a sentence in the out-of-sample contains \SI{0.09}{\percent} more negations than the second half. The latter is also statistically significant at the \SI{0.1}{\percent} level.

\subsection{Robustness Checks}
\label{sec:robustness_checks}

\begin{figure}
 		\centering
 		\includegraphics[width=\linewidth]{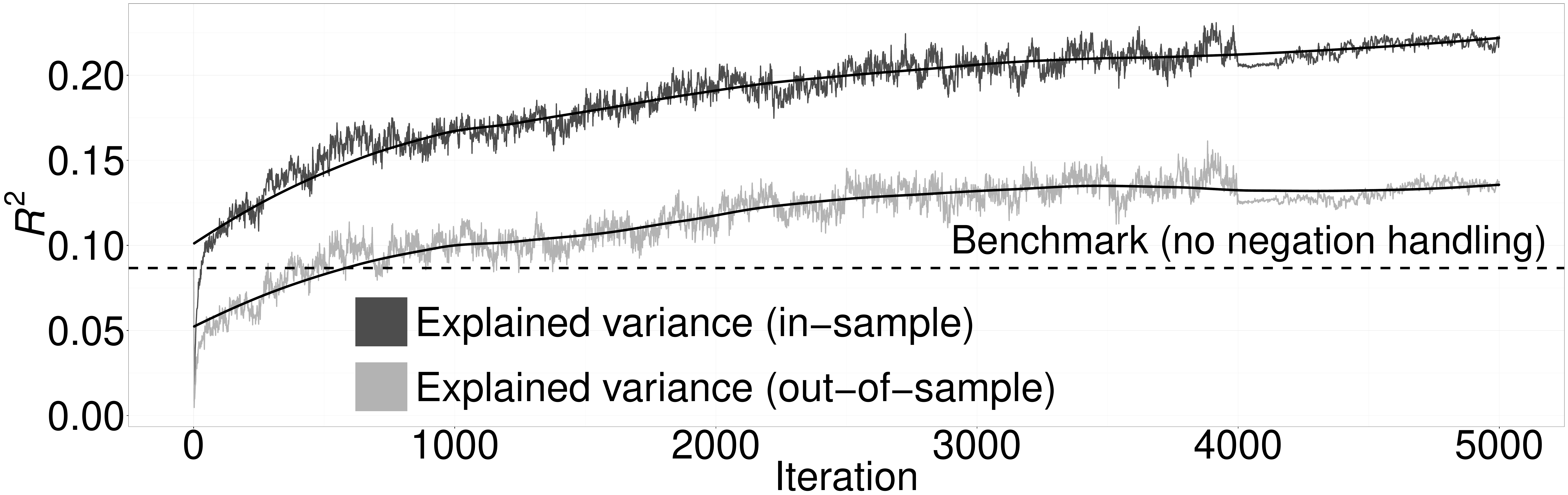}
 		\caption{The fluctuating series show the converging explained variance of user ratings based on tone (\ie sentiment) across different learning iterations using 10-fold cross validation, while the uniform lines shows it after smoothing. Here, the dark gray series corresponds to the in-sample set, whereas the light gray series corresponds to the out-of-sample set. The horizontal line denotes the \boldmath{$R^2$} in the benchmark setting (without handling negations).}
 		\label{fig:performance_correlation}
 	\end{figure}

We investigate the convergence of the reinforcement learning process to a stationary policy. Accordingly, \Cref{fig:performance_correlation} visualizes the proportion of variance of the gold standard that is explained by the tone for the first \num{5000} learning iterations. 
Here, the horizontal lines denote the explained variance in the benchmark setting (\ie no negation handling). Both the in-sample and out-of-sample {$R^2$} improve relatively quickly and outperform the benchmark considerably. In the end, we use our above policy based on \num{4000} iterations, since the next \num{1000} iterations consistently show fluctuations below \SI{0.05}{\percent} in terms of in-sample {$R^2$}. This pattern indicates a fairly stationary outcome. 


Next, we compare the performance of our reinforcement learning approach to common rules proposed in the literature~\citep{Hogenboom.2011,Taboada.2011}, which essentially try to imitate the grammatical structure of a sentence. For this purpose, the negation rules search for the occurrence of specific cues based on pre-defined lists and then invert the meaning of a fixed number of surrounding words. Hence, we apply the individual rules to each document and again compare the out-of-sample $R^2$; see \Cref{tbl:results_rules} for results. Negating a fixed window of the next \num{4} words achieves the highest fit among all rules similar to~\citep{Dadvar.2011}. This rule exceeds the benchmark with no negation handling by \SI{10.84}{\percent}. Most importantly, our approach works even more accurately, and dominates all of the rules, outperforming them by at least \SI{43.39}{\percent}. 

\begin{table}
\vspace{0pt}
\centering
\sisetup{round-mode=places,round-precision=4}
\renewcommand{\arraystretch}{1.3}
\tiny
	\begin{tabular}{lS}
		\toprule
			\textbf{Approach} & \textbf{Correlation}\\
			\midrule
			Benchmark: no negation handling & 0.08668691\\
			\midrule
			Negating all subsequent words & 0.08397561 \\
			Negating the whole sentence & 0.06608021\\
			Negating a fixed window of 1 word & 0.09184047\\
			Negating a fixed window of 2 words &  0.09483562\\
			Negating a fixed window of 3 words & 0.09527161\\
			Negating a fixed window of 4 words & 0.09624719\\
			Negating a fixed window of 5 words & 0.09614343\\
			\midrule
			Our approach (based on reinforcement learning) & 0.1377806\\	
			\bottomrule
	\end{tabular}
\caption{Table compares the out-of-sample $R^2$ (using 10-fold cross validation) of our approach and different rules from previous literature.}
\label{tbl:results_rules}
\end{table}

Finally, we evaluated further setups and methods for handling negations. We first tested alternative action sets for reinforcement learning that not only negate single words but also whole phrases, including backward negations. However, this configuration leads to inferior $R^2$ values on both the in-sample and out-of-sample set. We also explored other dictionaries of opinionated words and the performance of generative probabilistic models. Here, we find similar results for alternative dictionaries but inferior results for generative probabilistic models. All results confirm our findings.\footnote{Available on request.} 

\subsection{Comparison with Negation Processing in Financial News}

Our second case study investigates information processing in financial markets by analyzing how qualitative content in financial disclosures influences stock prices. For this purpose, we use \num{14463} regulated ad~hoc announcements\footnote{Kindly provided by the Deutsche Gesellschaft f{\"u}r Ad-Hoc-Publizit{\"a}t~(DGAP).} from European companies, all of which are written in English. These entail several advantages; for example, ad~hoc announcements must be authorized by company executives, their content is largely quality-checked by federal authorities and previous evidence finds a strong relationship between content and subsequent stock market reaction~\citep{Muntermann.2007}. As our gold standard, we calculate the daily abnormal return of the corresponding stock~\citep{Konchitchki.2011,MacKinlay.1997}. We measure the tone in these disclosures with the help of a finance-specific dictionary, the Loughran and McDonald dictionary~\citep{Loughran.2011}. This dictionary contains \num{354} entries with positive polarity, as well as \num{2350} entries marked as negative.

As detailed below, we derive a policy for negation processing and briefly introduce our main findings. Again, our reinforcement learning approach improves the link between tone and market response. Our benchmark without negation handling yields an out-of-sample {$R^2$ of \num{0.0042}}, while our method increases this by {\SI{37.02}{\percent}}, resulting in an out-of-sample {$R^2$} for 10-fold cross validation of {\num{0.0057}}. Both the absolute {$R^2$} and its improvements are -- as expected -- higher for movie reviews; this is a domain-specific disparity since \emph{\enquote{very few control variables predict next-day returns}} in efficient markets~\citep{Tetlock.2008}.

Next, we apply the learned policy to the out-of-sample documents and record its effects. Interestingly, we find that negations are less frequent in financial news. On average, each document contains \num{5.30} separate negation scopes that invert \SI{10.88}{\percent} of all opinion words. The length of the corresponding negation scopes is shorter, ranging from 1 to 14 words with an average length of \num{1.60} words. Similarly, to the results for the movie reviews, we find that the end of a document is more likely to contain negations than the beginning. On average, the second half of an out-of-sample announcement contains \SI{6.45}{\percent} more negation that its first half. It is noteworthy that this difference is significant at the 10\,\% level using a two-sided Welch $t$-test. This might suggest that authors of financial disclosures utilize negations as a tool to convey negative information through positive words. Overall, the results show that negations are domain-specific and depend on the particularities of the chosen prose. Additionally, the comparison strongly affirms that negation handling enhances the understanding of information processing for natural language of an arbitrary domain.

\section{Discussion}
\label{sec:discussion}

In the following sections, we discuss the implications of our research, as our method not only improves text comprehension, but also suggests a new approach to understanding decision-making in the social and behavioral sciences. Furthermore, our research is highly relevant for practitioners when extending information systems with interfaces for natural language. 

\subsection{Extensibility}
\label{sec:extensibility}

Our method of negation learning is not limited to the study of tone or sentiment; on the contrary, one can easily adapt it to all applications of natural language processing which utilize a gold standard and where negations play an important role. To accomplish this, one replaces the function calculating the sentiment with a corresponding counterpart that maps words onto a gold standard for the given application. Our only requirement is that this function takes into consideration -- in some way -- whether each word is negated or not. 

To better illustrate this concept, we briefly describe how this works using two examples. We first consider a medical question-answering system into which users enter their symptoms and, in return, are provided a list of potential illnesses. The system bases its answers on a collection of medical reports and one measures its performance by counting the number of correctly retrieved answers relative to the given input. For example, the system should return \emph{\textquote{fever}} for input \emph{\textquote{flu}} when the corpus contains \emph{\textquote{a flu causes fever}}. Reinforcement learning can improve accuracy in the presence of negations; \ie it learns that diseases can also be unrelated to symptoms, as in the statement, \emph{\textquote{A flu does not result in high blood pressure}}. As a second example, we assume an information system for negotiations that proposes offers to customers, who then reply in natural language. Subsequently, the information system automatically determines whether a customer's response was positive or negative based on its content. A na{\"i}ve bag-of-words model considers only specific cues (such as \emph{accept} without context), whereas our approach can even learn to correctly classify cases with negations, \eg by appending the prefix \emph{\textquote{not\_}} to words that are negated~\citep{Wiegand.2010}.

We now generalize the reward function in order to search for an optimal negation policy for the above applications. For each document $d$ with gold standard $y_d$, we calculate the predictive performance $\mathit{perf}_d^0$ that should forecast the gold standard negation handling, as well as $\mathit{perf}_d^\pi$ using the current policy $\pi$. The agent then gains a reward
\begin{equation}
r_i = \begin{cases}
0 , & \text{if } a_i = \text{Negated and } i < N_d , \\
c , & \text{if } a_i = \neg\text{Negated and } i < N_d , \\
\abs{y_d - \mathit{perf}_d^0} - \abs{y_d - \mathit{perf}_d^\pi} , & \text{if } i = N_d
\end{cases}
\end{equation}
with a suitable constant $c$. The first two cases add a small reward $c$ for default (\ie non-negating) actions to avoid overfitting, while the last case rewards how much better the current policy approximates the gold standard compared to no treatment of negations. This definition thus extends reinforcement learning to seek optimal negation processing across almost arbitrary applications of natural language processing.

\subsection{Limitations}

The current research faces a number of limitations, which can provide investigative possibilities for further works as follows: first and foremost, our method exhibits shortcomings when language is intricate, such as when piece of text refers to content that is located in an entirely different part of the document. Sometimes one even requires additional background knowledge to correctly interpret the content, as in the statement, \emph{\textquote{The movie was not at all different from the last one}}. This complexity poses challenges to natural language processing -- not only for our method but also for those discussed in the related work. In addition, we predominantly focus on implementations where negations have a forward-looking scope, \ie a negation cue affects subsequent words but not words that precede it. Therefore, we have also tested variants with a backward-looking analysis as part of our robustness checks; however, this offers opportunities for additional variations with advanced actions which could, for instance, invert the meaning of the full sentence or the subsequent object in order to further improve accuracy. Finally, further effort is necessary to develop an unsupervised variant which eliminates the need for a gold standard. 

\subsection{Implications for IS Research}

The unique and enduring purpose of IS research as an academic field is to understand and improve the ways in which people, organizations and businesses create value with information~\citep{Briggs.2015,Nunamaker.2011}. Hence, the design and implementation of systems to provide \emph{\textquote{the right information to the right person at the right time was the raison d'{\^e}tre of early IS research}}~\citep{Aggarwal.2012}. In the past, \textquote{information} predominantly referred to structured data, while companies nowadays also exploit unstructured data and especially textual materials. This development has found its way into IS research, which thus focuses on how textual information is processed. Among the earliest references to this area of inquiry is an article in \emph{Management of Information Systems Quarterly} from 1982 that explicitly addresses information processing~\citep{Robey.1982}. 

The field of information processing has gained great traction with advances in neuroscience and NeuroIS~\citep{Dimoka.2011}. By acquiring neuro-physiological data, scholars can gather information on how the human brain reacts to external stimuli. For this purpose, one measures (neuro-)physiological parameters (\eg heart rate and skin conductance) to study the information processing of human agents~\citep{Riedl.2014}. This makes it possible to measure informational and cognitive overload in users in the course of their interaction with information systems~\citep{Riedl.2014} and text-based information~\citep{Minas.2014}. However, NeuroIS remains very costly and the methods exhibit many weaknesses~\citep{Dimoka.2012}. For example, interpreting data from functional magnetic resonance imaging~(fMRI) is hampered by the complexity and non-localizable activities of the brain.

Our computational intelligence method promises to fill the gap in existing approaches to understanding negations. It is analogous to revealed preferences estimations in economics, where the choices of individuals reveal the individuals' latent utility function, since we utilize text documents that are tagged by the users with a rating or gold standard. In addition, applying computational intelligence offers the potential to automatically unveil negations in texts -- without the need to manually label individual words. This entails several advantages as the understanding of language is highly subjective and, in contrast, we derive the (latent) negation model that best fits the data. The results can thus also contribute to linguistic and psychological models of negation usage and representation~\citep{Khemlani.2012}. Overall, reinforcement learning manifests immense potential for future IS research involving the study of information processing in depth.

\subsection{Implications for Practitioners}


Previous IS research argues that \emph{\textquote{a basic issue in the design of expert systems is how to equip them with representational and computational capabilities}} \citep{Zhang.1987}. As a remedy, this paper presents a tool to practitioners in order to improve the automated processing of natural language in their information systems. As such, our methodology can enhance the accuracy of decision support based on textual data. It does not necessarily require changes in the derivation of the original algorithms; instead, our methodology can be built on top of routines and thus enables a seamless integration into an existing tool chain.

Practitioners can benefit from negation handling when assessing the semantic orientation of written materials. For example, in the case of recommender systems and opinion mining, texts provide decision support by tracking the public mood in order to measure brand perception or judge the launch of a new product based on blog posts, comments, reviews or tweets. Based on our case study, we see a significant improvement of up to \SI{58.94}{\percent} in explained variance by adjusting for negated text units.

A better understanding of human language can spark business innovations in multiple areas. For instance, our approach facilitates the interactive control of information systems through natural language, such as in question-answering systems. With the advent of cognitive computing, the accurate processing of natural language will gain even more in importance~\citep{Modha.2011}. Ultimately, the relevance of our methodology goes beyond these examples and comprises almost all text-based applications of individuals, organizations and businesses.

\section{Conclusion}

Information is at the heart of all decision-making that affects humans, businesses and organizations. Consequently, understanding the formation of decisions represents a compelling research topic and yet knowledge gaps become visible when it comes to information processing with regard to natural language. Negations, for example, are a frequently utilized linguistic tool for expressing disapproval or framing negative content with positive words; however, existing methods struggle to accurately recognize and interpret negations in written text. Moreover, these methods are often tethered to large volumes of manually labeled data, which introduce an additional source of subjectivity and noise. 

In order to address these shortcomings, this paper develops a novel approach based on reinforcement learning, which has the advantage of being human-like and thus capable of learning to replicate human decision-making. As a result, our evaluation shows superior performance in predicting negation scopes, while this method also reveals an unbiased approach to identifying negation scopes based on an exogenous response variable collected at document level. It thereby sheds light on the \textquote{ground truth} of negation scopes, which would have otherwise been latent and unobservable. In addition, reinforcement learning allows for hypothesis testing in order to pinpoint how humans process and act on negations. For instance, this paper demonstrates that negations are unequally distributed across document segments, showing that the second half of movie reviews and financial news items contain significantly more negations than the first half. Our approach serves as an intriguing alternative or supplement to experimental research, as it unleashes computational intelligence for the purpose of performing behavioral research, thereby fostering unprecedented insights into human information processing.


\bibliographystyle{model1-num-names}
\bibliography{literature}

\end{document}